\documentclass{ws-procs9x6}
\usepackage{todo}
\usepackage{hyperref}
\usepackage{siunitx}
\usepackage{xargs}
\usepackage[pdftex,dvipsnames]{xcolor}  
\newcommandx{\unsure}[2][1=]{\todo[linecolor=red,backgroundcolor=red!25,bordercolor=red,#1]{#2}}
\newcommandx{\towrite}[2][1=]{\todo[linecolor=Cyan,backgroundcolor=Cyan!25,bordercolor=Cyan,#1]{#2}}
\newcommandx{\change}[2][1=]{\todo[linecolor=blue,backgroundcolor=blue!25,bordercolor=blue,#1]{#2}}
\newcommandx{\info}[2][1=]{\todo[linecolor=OliveGreen,backgroundcolor=OliveGreen!25,bordercolor=OliveGreen,#1]{#2}}
\newcommandx{\improve}[2][1=]{\todo[linecolor=Plum,backgroundcolor=Plum!25,bordercolor=Plum,#1]{#2}}
\newcommandx{\thiswillnotshow}[2][1=]{\todo[disable,#1]{#2}}
\usepackage{acronym}
\acrodefplural{CNN}[CNNs]{convolutional neural networks}
 \newacro{ADR}{Automatic domain randomization}
 \newacro{AGV}{automated guided vehicle}
 \newacro{ANN}{artificial neural network}
 \newacro{CNN}{convolutional neural Network}
 \newacro{cobots}{collaborative robots}
 \newacro{DoF}{degrees of freedom}
 \newacro{FDM}{fused deposition modeling}
 \newacro{FFM}{fused filament fabrication}
 \newacro{LSTM}{long short-term memory}
 \newacro{MAML}{Model-Agnostic Meta-Learning}
 \newacro{ML}{meta-learning}
 \newacro{RL}{reinforcement learning}
\begin{document}
%
\title{Towards Intelligent Pick and Place Assembly of Individualized Products Using Reinforcement Learning}
\author{Caterina Neef$^{1,*}$, Dario Luipers$^{1,2,*}$, Jan Bollenbacher$^3$, Christian Gebel$^1$ and Anja Richert$^{1,*}$}
\address{$^1$Cologne Cobots Lab\footnote[2]{The Cologne Cobots Lab is an interdisciplinary research lab of the TH Köln – University of Applied Sciences, with its main research focus on collaborative and social robotics.}, $^2$Laboratory of Manufacturing Systems, $^3$Institute of Telecommunications Technology, TH Köln - University of Applied Sciences,\\
Betzdorfer Str. 2, 50679 Cologne, Germany\\
$^*$E-mail: \{caterina.neef,dario.luipers,anja.richert\}@th-koeln.de\\
www.th-koeln.de}
\begin{abstract}
%
Individualized manufacturing is becoming an important approach as a means to fulfill increasingly diverse and specific consumer requirements and expectations.
While there are various solutions to the implementation of the manufacturing process, such as additive manufacturing, the subsequent automated assembly remains a challenging task.
As an approach to this problem, we aim to teach a collaborative robot to successfully perform pick and place tasks by implementing reinforcement learning.
For the assembly of an individualized product in a constantly changing manufacturing environment, the simulated geometric and dynamic parameters will be varied.
Using reinforcement learning algorithms capable of meta-learning, the tasks will first be trained in simulation.
They will then be performed in a real-world environment where new factors are introduced that were not simulated in training to confirm the robustness of the algorithms.
The robot will gain its input data from tactile sensors, area scan cameras, and 3D cameras used to generate heightmaps of the environment and the objects.
The selection of machine learning algorithms and hardware components as well as further research questions to realize the outlined production scenario are the results of the presented work.

\end{abstract}
\keywords{machine learning, reinforcement learning, meta-learning, individualized manufacturing, collaborative robotics}
\bodymatter
\section{Introduction}
%
For decades, robots have been used to automate tasks in the industry sector.
Conventional industrial robots are taught to perform one task at a time, are competent at executing this single task and can perform thousands of repetitions accurately.
While the automation of such tasks has led to an increase in efficiency and a decrease in manufacturing costs for mass production, it is less applicable for the individualized consumer expectations of today's economy.
Globalization, digitalization and the resulting growth of markets have led to an increasing number of product variants and shorter product life cycles \cite{brettel2014}.
Customers now demand highly individualized products that are designed specifically for them.
This change is observable in a wide range of industrial fields \cite{outon2019}.\par
The key challenge for individualized products is to avoid an increase in costs compared to established manufacturing approaches like mass production, even if factories are located in high-cost countries \cite{lanz2017}.
The advantages of individualized production are higher flexibility, fast response rates to customer decisions and a more efficient use of resources.
An example is the health-care sector, where personalized medicine is becoming increasingly important, and additive manufacturing is being used for the production of biomaterials, implants and prosthetics \cite{zadpoor2017}.
To enable individualized manufacturing, traditional programming of machines with repetitive tasks is no longer applicable \cite{outon2019}.
When environments and objects change, conventional robots are unable to perform assembly tasks with similar success rates.\par
We therefore propose to use \ac{RL} algorithms capable of \ac{ML} to enable robots to accomplish highly individualized pick and place tasks as an important part of the product assembly process.

\section{State of the Art: Machine Learning and Robotics}
\label{sec:sota}
Recently, \ac{RL} has achieved great success in a wide range of different tasks and complex games (e.g. the strategy board game Go \cite{Silver2017}).
The implementation of \ac{RL} and \ac{ML} seems promising to enable a robot to perform a pick and place task for unknown objectives and destinations.
In \ac{RL}, an agent interacts with the environment and receives its state.
Based on this state, the agent takes an action and receives a new state and reward for the chosen action.
Each \ac{RL} algorithm is designed specifically for a certain task in terms of its architecture and training.
A major drawback of this approach is the necessity to train the \ac{RL} agent from scratch for each task.\par
\ac{ML} is an approach to overcome this shortcoming by designing an algorithm in such a way that the agent learns how to learn from a broad distribution of similar tasks.
Similar to human learning, an \ac{ML} agent can apply knowledge it has gained from previously solved corresponding tasks to learn a new task with only a small amount of data.\par
Recent \ac{ML} algorithms suitable for \ac{RL} can be divided into two categories depending on their architecture and optimization goals:
\begin{enumerate}
	\item Model-based meta-learning approaches \cite{Duan2016,Mishra2017} generalize to a wide range of learning scenarios, seeking to recognize the task identity from a few data samples and adapting to the tasks by adjusting a model's state (e.g. \ac{LSTM} internal states)
	\item \ac{MAML} \cite{Finn2017} seeks an initialization of model parameters such that a small number of gradient updates will lead to fast learning on a new task, offering flexibility in the choice of models
\end{enumerate}
A promising approach in which machine learning without \ac{RL} and \ac{ML} is used for pick and place tasks is described in \citenum{zakka2019}:
A six \ac{DoF} UR5e robot (Universal Robots, Odense, Denmark) with a suction module is used to perform the tasks, and a camera generates a 3D heightmap as input data.
Using \acp{CNN}, a correspondence between an object surface and the related placement location is generated.\par
Another relevant RL approach is introduced by OpenAI, who have trained a robotic hand to solve a Rubik's Cube despite external perturbations \cite{openai2019}.
The main points of this approach are:
\begin{itemize}
  \item An actor-critic consisting of an \ac{ANN} equipped with \ac{LSTM} cells to install internal memory
  \item \ac{ADR} to generate diverse environments with randomized physics and dynamics (e.g. weight and size of the manipulated object)
\end{itemize}
This results in a system with high robustness and high success rates in the transfer from simulation to testing in the real-world environment.
Due to the combination of internal memory and \ac{ADR}, this approach also shows signs of emerging \ac{ML}.\par
In \ac{RL}, it is necessary to provide the learning agent with an extrinsic reward signal.
This enables the agent to determine if the actions applied to the environment have a positive effect in the long run.
Extrinsic reward signals are called sparse if the reward for a certain action is temporally disentangled from the reward, e.g. only a positive reward is given after every successful task.
To tackle this problem of sparse extrinsic rewards, we can divide the approaches in literature into two classes.
First, by changing the reward function, e.g. using curiosity-driven exploration \cite{PathakAED17}, which introduces an intrinsic reward function.
This function encourages the agent to experience novel states.
Second, hierarchical \ac{RL} methods which try to divide the main task into a sequence of sub-goals can be used.
While the main goal is to successfully perform the task, the agent first learns to find a policy for the sub-goals.
One popular candidate for this are FeUdal Networks \cite{VezhnevetsOSHJS17}, in which the agent is split into two parts.
The manager learns to formulate goals and the worker is intrinsically rewarded to follow the goal.
A similar approach is Hierarchical Actor-Critic \cite{levy2017learning}, in which the agent learns to set sub-goals to reach the main goal.
This is achieved by extending the idea of Hindsight Experience Replay \cite{AndrychowiczWRS17} to the hierarchical setting by establishing goals a fixed number of low-level actions away from the previous state.
Multiple policies can be learned independently.
On a sub-goal level, the focus is learning the sequences of sub-goal states which can reach the main goal state.
To achieve these sub-goal states, the lower-level policies learn the low-level action sequences.\par

\section{Concept for Manufacturing Scenario}
\label{sec:scenario}
\begin{figure}[h!]
    \centering
    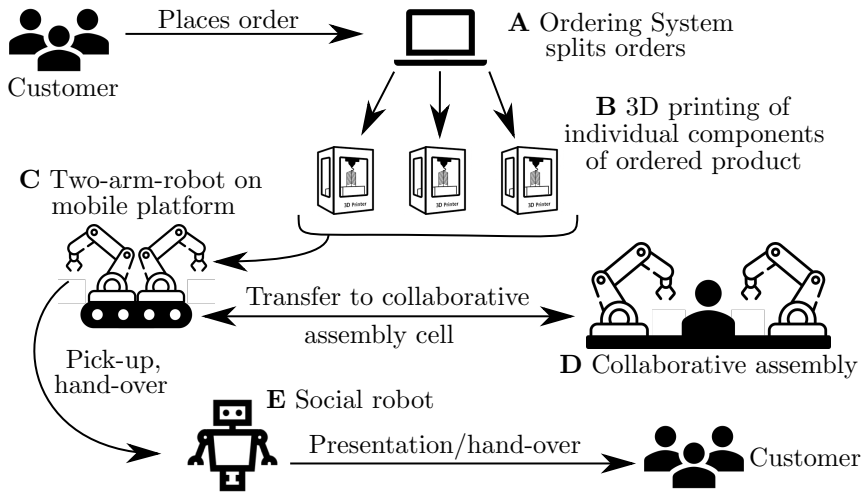
    \caption{A customer places an order for an individualized product using the ordering system (A).
    This system splits the order into the individual components, which are then manufactured using additive manufacturing (B).
    Once the manufacturing process is completed, a two-arm-robot, mounted on an \ac{AGV} (C), removes the components from the 3D printers and transports them to the collaborative assembly cell (D).
    This assembly cell consists of one or more \ac{cobots} and one or more human workers, collaboratively assembling the individual components into the final product.
    The number of \ac{cobots} and human workers can be adjusted based on need for the specific task.
    Once the assembly process is completed, the two-arm-robot transports the finished product and hands it over to a social robot (E), which presents and hands over the product to the customer.}
    \label{fig:scenario}
\end{figure}
\begin{figure}[h!]
    \centering
    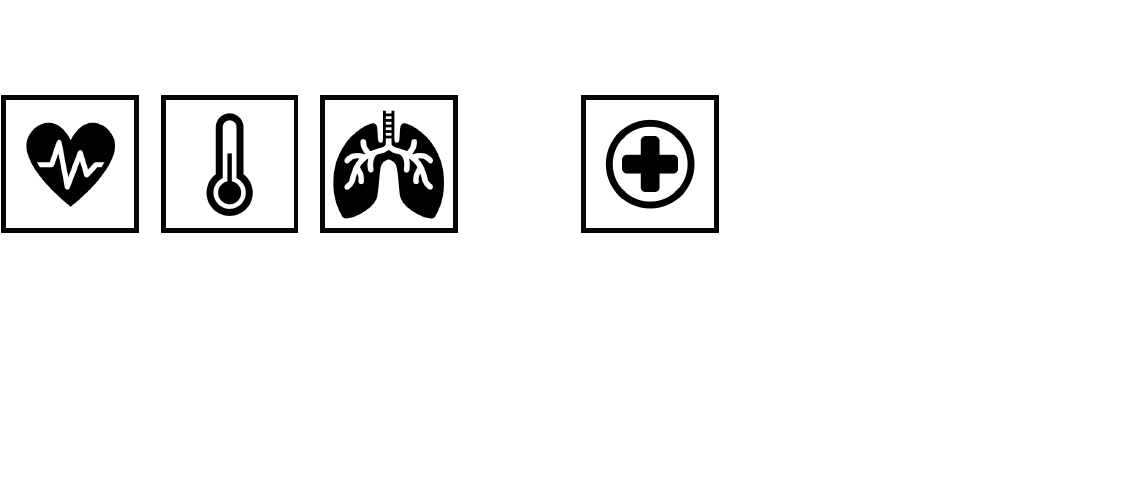
    \caption{The goal of the assembly task is to pick individual health care sensors, (e.g. heart rate (HR), body temperature (T), blood oxygen saturation (SpO\textsubscript{2})), place them into their cases, then place the sensor cases in a health care sensor station for the monitoring of health conditions. As the sensor combination changes with each user, \ac{RL} will be used to successfully train an agent to perform each individual assembly task.}
    \label{fig:sensors}
\end{figure}
The development of intelligent pick and place tasks for the assembly of products using \ac{RL} is an integral part of the manufacturing scenario we are setting up in the Cologne Cobots Lab.
It combines individualized production using additive manufacturing, autonomous mobile systems that transport components, as well as collaborative and social robotics.
The complete scenario is shown in \autoref{fig:scenario}.\par
As described in \autoref{sec:sota}, we aim to perform object manipulation tasks using \ac{RL} in a real-world scenario, in which a specific and useful product is manufactured and assembled.
Our goal is to assemble individualized sensor cases for different health care sensors (e.g. to measure body temperature, heart rate or blood oxygen saturation), as shown in \autoref{fig:sensors}.
These cases will be used in our research concerned with social robots conducting health assessments \cite{richert2019}.
This is a practical product for research in highly flexible areas, such as the manufacturing and health care sectors.
It is also a well-suited demonstrator for the application of \ac{RL} algorithms.
The assembly of individualized products is desirable, as different users with different health conditions require different kinds of information.
Therefore, the combination of sensors can be adapted for each individual user and the assembly process will differ for each new product.
The long-term goal regarding these sensors is to create individualized wearable devices with different kinds and numbers of health care sensors.
These parameters offer a promising approach to a hybrid job shop scheduling or action planning system in which human and robot actions are combined in an optimized way.

\section{Approach: Hardware and Machine Learning}
In our manufacturing scenario, we will be using various hardware and software/machine learning components, described in the following.\par
The individual sensor cases will be manufactured using \ac{FDM}/\ac{FFM}.
For the assembly, we will use a collaborative robotic arm that meets safety standards defined in ISO/TS 15066:2016 \cite{standard2016iso}.
The arm also has a high pose repeatability, at least six \ac{DoF}, and a payload of \SI{>= 0.5}{\kilogram} for fine and \SI{>= 3}{\kilogram} for gross manipulation tasks.
To receive additional feedback during object manipulation tasks, its gripper will be outfitted with tactile sensors.
This will improve the efficiency and robustness of the grasping task.
The sensor provides feedback regarding the grip quality confidence and enables slip detection of the grasped object, which can then be counteracted by improving the applied force of the gripper on the object.
For the object detection, we will use a 3D scanner to create heightmaps of the objects.
Several area scan cameras will be implemented for vision-based information from various angles and to determine the orientation of the objects.\par
In order to successfully teach the robot to perform the pick and place task, the trained machine learning algorithm needs to recognize which produced element belongs to the corresponding case.
To accomplish this, the tools presented in \autoref{sec:sota} will be implemented and combined.
The heightmaps applied by Google \cite{zakka2019} will be used as input for the \ac{RL} agent.
By implementing \ac{ADR} \cite{openai2019}, the agent will be trained to realize a robust system with a high success rate in the transfer from simulation to the real world.
Additionally, due to the implementation of \ac{ML} and solutions for sparse reward, the learning time of the agent will be decreased.\par
\section{Conclusion and Outlook}
In this paper, we propose an approach to successfully perform intelligent pick and place tasks for the assembly of individualized products using \ac{RL} algorithms capable of \ac{ML}.
A combination of the algorithms presented in this work will be implemented to develop an autonomous robotic system capable of performing these tasks.
With a combination of \ac{RL}, \ac{ML}, \ac{ADR} and other machine learning tools, problems like sparse reward or transfer learning can be solved.
The pick and place tasks are first performed in simulation, then in a real-world environment using a collaborative robot, equipped with tactile sensors, area scan cameras and 3D cameras.
This demonstrator will then be used to study and answer the following research questions, which are both of technical and socio-technical nature:
\begin{itemize}
    \item How can we apply (a combination of) machine learning algorithms to generalize pick and place assembly tasks (i.e. various weights, sizes, geometries, quantities) for individualized products?
    \item Which algorithms have which impact on the robustness of the system? How can we assure that the robustness reached in simulation can be transferred to the real world environment?
    \item  How can we implement a dynamic work space for the robot when working collaboratively with a human? How can a human be integrated into the collaborative assembly process in a way that is both sensible and effective?
\end{itemize}
In the future, we plan on fully implementing the developed assembly process into our manufacturing scenario described in \autoref{sec:scenario}.
The manufacturing scenario includes the transportation of individual parts using \acp{AGV} and presenting the final product to the customer.
A further goal is to study the collaborative assembly process between humans and robots.
This is the focus of another research project in our lab, which aims to achieve adaptive human-robot collaboration through the implementation of sensors to detect the user's status (e.g. focus, stress).
The combined results of these projects will contribute to an optimal collaborative working process.\par
%
%

%
\todos
\bibliographystyle{ws-procs9x6}
\bibliography{bibliography}

\begin{thebibliography}{10}

\bibitem{brettel2014}
M.~Brettel, N.~Friederichsen, M.~Keller and M.~Rosenberg, How {V}irtualization,
  {D}ecentralization and {N}etwork {B}uilding {C}hange the {M}anufacturing
  {L}andscape: {A}n {I}ndustry 4.0 {P}erspective, {\em International {J}ournal
  of {M}echanical, {I}ndustrial {S}cience and {E}ngineering} {\bf 8}, 37
  (2014).

\bibitem{outon2019}
J.~L. Out{\'o}n, I.~Villaverde, H.~Herrero, U.~Esnaola and B.~Sierra,
  Innovative {M}obile {M}anipulator {S}olution for {M}odern {F}lexible
  {M}anufacturing {P}rocesses, {\em Sensors} {\bf 19}, p. 5414  (2019).

\bibitem{lanz2017}
M.~Lanz and R.~Tuokko, Concepts, {M}ethods and {T}ools for {I}ndividualized
  {P}roduction, {\em Production Engineering} {\bf 11}, 205  (2017).

\bibitem{zadpoor2017}
A.~A. Zadpoor and J.~Malda, Additive {M}anufacturing of {B}iomaterials,
  {T}issues, and {O}rgans  (2017).

\bibitem{Silver2017}
D.~Silver, J.~Schrittwieser, K.~Simonyan, I.~Antonoglou, A.~Huang, A.~Guez,
  T.~Hubert, L.~Baker, M.~Lai, A.~Bolton {\em et~al.}, Mastering the game of go
  without human knowledge, {\em Nature} {\bf 550}, 354  (2017).

\bibitem{Duan2016}
Y.~Duan, J.~Schulman, X.~Chen, P.~L. Bartlett, I.~Sutskever and P.~Abbeel,
  R{L}\textsuperscript{2}: {F}ast {R}einforcement {L}earning via {S}low
  {R}einforcement {L}earning, {\em arXiv preprint arXiv:1611.02779}   (2016).

\bibitem{Mishra2017}
N.~Mishra, M.~Rohaninejad, X.~Chen and P.~Abbeel, A {S}imple {N}eural
  {A}ttentive {M}eta-{L}earner, {\em arXiv preprint arXiv:1707.03141}   (2017).

\bibitem{Finn2017}
C.~Finn, P.~Abbeel and S.~Levine, Model-{A}gnostic {M}eta-{L}earning for {F}ast
  {A}daptation of {D}eep {N}etworks, in {\em Proceedings of the 34th
  International Conference on Machine Learning-Volume 70\/}, 2017.

\bibitem{zakka2019}
K.~Zakka, A.~Zeng, J.~Lee and S.~Song, Form2{F}it: {L}earning {S}hape {P}riors
  for {G}eneralizable {A}ssembly from {D}isassembly, {\em arXiv preprint
  arXiv:1910.13675}   (2019).

\bibitem{openai2019}
OpenAI, I.~Akkaya, M.~Andrychowicz, M.~Chociej, M.~Litwin, B.~McGrew,
  A.~Petron, A.~Paino, M.~Plappert, G.~Powell, R.~Ribas, J.~Schneider,
  N.~Tezak, J.~Tworek, P.~Welinder, L.~Weng, Q.~Yuan, W.~Zaremba and L.~Zhang,
  Solving {R}ubik's {C}ube with a {R}obot {H}and, {\em arXiv preprint}
  (2019).

\bibitem{PathakAED17}
D.~Pathak, P.~Agrawal, A.~A. Efros and T.~Darrell, Curiosity-driven exploration
  by self-supervised prediction, in {\em Proceedings of the IEEE Conference on
  Computer Vision and Pattern Recognition Workshops\/}, 2017.

\bibitem{VezhnevetsOSHJS17}
A.~S. Vezhnevets, S.~Osindero, T.~Schaul, N.~Heess, M.~Jaderberg, D.~Silver and
  K.~Kavukcuoglu, Fe{U}dal networks for hierarchical reinforcement learning, in
  {\em Proceedings of the 34th International Conference on Machine
  Learning-Volume 70\/}, 2017.

\bibitem{levy2017learning}
A.~Levy, G.~Konidaris, R.~Platt and K.~Saenko, Learning {M}ulti-{L}evel
  {H}ierarchies with {H}indsight, {\em arXiv preprint arXiv:1712.00948}
  (2017).

\bibitem{AndrychowiczWRS17}
M.~Andrychowicz, F.~Wolski, A.~Ray, J.~Schneider, R.~Fong, P.~Welinder,
  B.~McGrew, J.~Tobin, O.~P. Abbeel and W.~Zaremba, Hindsight {E}xperience
  {R}eplay, in {\em Advances in neural information processing systems\/}, 2017.

\bibitem{richert2019}
A.~Richert, M.~Schiffmann and C.~Yuan, A {N}ursing {R}obot for {S}ocial
  {I}nteractions and {H}ealth {A}ssessment, in {\em International Conference on
  Applied Human Factors and Ergonomics\/}, 2019.

\bibitem{standard2016iso}
S.~ISO, {ISO}/{TS} 15066: 2016 {R}obots and robotic devices -- {C}ollaborative
  robots  (2016).

\end{thebibliography}
\end{document}